# EchoPrime: Multi-Video View-Informed Vision-Language Model for Comprehensive Echocardiography Interpretation

Milos Vukadinovic[1,2], Xiu Tang[3], Neal Yuan[4], Paul Cheng[3], Debiao Li[5], Susan Cheng[1], Bryan He[6,*], David Ouyang[1,7,*]

1. Department of Cardiology, Smidt Heart Institute, Cedars-Sinai Medical Center, Los Angeles, CA
2. Department of Bioengineering, University of California Los Angeles, Los Angeles, CA
3. Division of Cardiology, Department of Medicine, Stanford University, Palo Alto, CA
4. Department of Medicine, University of California, San Francisco, CA; Division of Cardiology, San Francisco Veterans Affairs Medical Center, San Francisco, CA
5. Biomedical Imaging Research Institute, Cedars-Sinai Medical Center, Los Angeles, CA
6. Department of Computer Science, Stanford University, Stanford, CA
7. Division of Artificial Intelligence in Medicine, Cedars-Sinai Medical Center, Los Angeles, CA

* Equal Contribution

Correspondence: bryanhe@cs.stanford.edu, david.ouyang@cshs.org

**Abstract**

Echocardiography is the most widely used cardiac imaging modality, capturing ultrasound video data to assess cardiac structure and function. Artificial intelligence (AI) in echocardiography has the potential to streamline manual tasks and improve reproducibility and precision. However, most echocardiography AI models are single-view, single-task systems that do not synthesize complementary information from multiple views captured during a full exam, and thus lead to limited performance and scope of applications. To address this problem, we introduce EchoPrime, a multi-view, view-informed, video-based vision-language foundation model trained on over 12 million video-report pairs. EchoPrime uses contrastive learning to train a unified embedding model for all standard views in a comprehensive echocardiogram study with representation of both rare and common diseases and diagnoses. EchoPrime then utilizes view-classification and a view-informed anatomic attention model to weight video-specific interpretations that accurately maps the relationship between echocardiographic views and anatomical structures. With retrieval-augmented interpretation, EchoPrime integrates information from all echocardiogram videos in a comprehensive study and performs holistic comprehensive clinical echocardiography interpretation. In datasets from two independent healthcare systems, EchoPrime achieves state-of-the art performance on 23 diverse benchmarks of cardiac form and function, surpassing the performance of both task-specific approaches and prior foundation models. Following rigorous clinical evaluation, EchoPrime can assist physicians in the automated preliminary assessment of comprehensive echocardiography.

**Main Text**

In recent years, medical artificial intelligence (AI) has progressed significantly, driven by the fast improvements of deep learning methods and the use of larger and larger medical datasets. AI has matched or surpassed the accuracy of clinical experts in various applications[1], such as skin cancer classification[2] and breast mammogram lesion detection[3]. However, because these models are tailored to specific tasks through supervised learning, they lack the ability to synthesize information in a holistic manner, unlike clinical experts that integrate multiple data points for a comprehensive assessment. Moreover, with thousands of possible diagnoses and diseases, it is impractical to train separate models for every individual medical task. This limitation gave rise to foundation models[4–12], task-agnostic models pre-trained on large datasets that demonstrate robust performance in a variety of downstream tasks. Foundation models have already been utilized across various medical subfields, including pathology[13], drug repurposing[14], chest X-ray[15], and retinal imaging[16].

Echocardiography, or cardiac ultrasound, is the most common form of cardiac imaging and benefits from high volume, low cost, portability, and lack of ionizing radiation. With the highest temporal resolution across all imaging modalities, echocardiography videos capture changes in heart motion and structure associated with cardiomyopathy, valvular disorders, tamponade, and arrhythmias. Progress towards a foundation model for echocardiography was made with EchoCLIP[17], which was trained on over 1 million echocardiogram videos and demonstrated good performance across a diverse range of benchmarks for cardiac image interpretation. However, EchoCLIP utilizes only a static image encoder from a single echocardiogram view, rather than incorporating available dynamic videos from a comprehensive ultrasound exam. As a result, it might miss key temporal and functional insights that are vital for echocardiographic analysis.

In this work, we introduce EchoPrime, a video-based foundation model for echocardiography, trained with contrastive learning on more than 12 million videos paired with expert interpretations and designed to synthesize data from multiple videos to deliver comprehensive interpretations. We test our model's performance on datasets across two healthcare systems on an extensive benchmark of multi-modal retrieval metrics, clinical echocardiography interpretation tasks covering all cardiac structures, and transfer learning tasks related to cardiac pathophysiology. We find that EchoPrime consistently outperforms other medical foundation models (BioMedCLIP and EchoCLIP) and either matches or exceeds the performance of task-specific ultrasound models. In additional analyses, we find that EchoPrime prioritizes views similar to clinical experts when assessing many echocardiogram videos through multiple instance attention. EchoPrime is the largest existing echocardiography AI model trained on over ten times the data of prior models and natively provides multi-view, multi-task, and multi-video assessments. To advance AI in medicine research, we publicly release code, weights, and a hosted demo.

## Results

EchoPrime is a video-based vision-language model trained with 12,124,168 echocardiography videos and paired text reports from 275,442 studies across 108,913 patients at Cedars-Sinai Medical Center (CSMC) (Table 1). EchoPrime consists of multiple modules integral for the interpretation of echocardiography, including a video encoder, text encoder, view classifier, and anatomical attention module (Figure 1). The video and text encoders are trained contrastively on sampled echocardiogram video clips and corresponding cardiologist report texts to learn a joint video-text representation space. A view classification model was trained on 77,426 sonographer labeled videos to classify B-mode and color Doppler videos into 58 standard echocardiographic views and utilized by an anatomical attention module to determine the relative importance of each echocardiogram video for interpretation tasks using multiple instance learning. During inference, EchoPrime provides a comprehensive interpretation of an echo study by assigning views to each video, mapping the videos into a contrastive joint video-text latent space, and finally retrieving study interpretations guided by the anatomical attention module.

## Automated echocardiogram interpretation without supervised learning

EchoPrime can simultaneously interpret a wide range of cardiac features and diagnoses that represent a wide range of pathophysiology seen over a decade of echocardiography at a large tertiary care center. To obtain echocardiogram exam interpretations, we employ retrieval augmented interpretation (RAI), a method similar to retrieval augmented generation (RAG)[18]. RAI works by retrieving the historical echocardiogram reports that best match input echocardiogram videos, and then filters the information from these reports into final output interpretations based on anatomical attention (see Methods for details). This approach allows a single foundation model to be applied directly to all echocardiography interpretation tasks, enabling the prediction of any feature present in the echocardiography report.

On a held out internal validation cohort from CSMC and an external validation cohort from Stanford Healthcare (SHC), EchoPrime outperformed both previously developed foundation models and task-specific echocardiography AI models (Table 2). On a diverse set of benchmarks for cardiac structure and function, EchoPrime had a mean AUC of 0.92 across 17 classification tasks, including tasks for which no prior echocardiography AI models as well as tasks with task-specific AI models. Without additional fine-tuning or training, EchoPrime outperforms or matches fully supervised models for a single-task prediction[19–24] including a 2% improvement over EchoNet-Dynamic[19] in R2 score for left systolic heart failure, 2% improvement in AUC over Echonet-TR[20] for assessing tricuspid regurgitation, 4% increase over EchoNet-MR[21] for assessing mitral regurgitation, and similar to improved performances for pericardial effusion[22] and aortic stenosis[23].

Our model also performed favorably when compared to other foundation models like BioMedCLIP[25], a general foundation model for biomedicine, and EchoCLIP[17], a prior foundation model for echocardiography (**Figure 2**B,C). EchoPrime demonstrated significant improvements over previous models in predicting features related to the motion of cardiac structures. Specifically, in estimating left ventricular systolic function, EchoPrime achieved a mean absolute error (MAE) of 4.8% on the internal dataset, compared to MAE of 26.9% and 7.0% for BioMedCLIP and EchoCLIP, respectively. On the external SHC dataset, EchoPrime demonstrated an MAE of 4.1%, while BioMedCLIP and EchoCLIP had MAE of 23.0% and 6.2%, respectively. In detecting aortic regurgitation, EchoPrime achieved an AUC of 0.88 versus 0.54 and 0.68 for BioMedCLIP and EchoCLIP in the internal dataset, and 0.89 externally versus 0.63 and 0.67 for BioMedCLIP and EchoCLIP in the external dataset (full results in Table 2) Our results show that EchoPrime outperforms other echocardiography and medical imaging foundation models, often by a significant margin, and matches or exceeds the performance of task-specific models.

**Leveraging multi-view echocardiographic data improves performance**

Our model synthesizes the full range of information from an echocardiogram study, including multiple videos of different views and clinical report text up to 512 tokens long. This marks a significant improvement over previous medical foundation models like BioMedCLIP and EchoCLIP, which processes only single view, individual images and handles text up to 77 tokens long. To test the impact of this design, we performed zero-shot cross-modal retrieval of videos-to-text and text-to-videos (

**Table *3*). On videos-to-text retrieval, the correct text report appeared in the top 10 matched text reports for 98% of the studies from the test set (Recall@10), outperforming EchoCLIP by 45%. Similarly, on the text-to-videos task our model achieved Recall@10 of 97%, outperforming EchoCLIP by 35%.

Cardiologists integrate information from multiple views and videos to provide a holistic assessment, and EchoPrime was designed to weight the interpretation of multiple videos from multiple views. To demonstrated the impact of integrating multiple clips, multiple videos, and multiple views, we compared the performance of the EchoPrime encoder when input is a single frame vs. single video vs. multiple videos vs. multiple videos with multiple instance attention weighting (Figure 2A). The gradual improvement across multiple tasks reflects the relevance of temporal and view-specific information to each task and the importance of the anatomic attention module (EchoPrime weighting). For each task, there was a continuous improvement in model performance as additional elements of EchoPrime design and modules were included.

**Anatomical attention provides interpretable weighting of key views**

Our view classifier achieves a one-vs-rest AUC of 0.997 for predicting 58 different standard echocardiographic views (Supplementary Figure 2A) on the internal test set consisting of 4,000 videos. The view prediction, as well as the underlying videos, were then used to train an anatomical attention module using a multiple instance attention framework to identify the most relevant views and videos for a given anatomic structure and report section (Figure 3). With the anatomical attention module, we enable EchoPrime to focus on the most informative views for the anatomy being evaluated and weigh conflicting assessments of different videos from the same study. (Supplementary Figure 2B).

For instance, in assessing the mitral valve, the model learned to focus on apical-2-chamber, apical-4-chamber, parasternal short axis on the mitral valve, and corresponding color Doppler views as the key videos. Similarly, the model identified that views where the IVC is clearly visible to be the most relevant for prediction tasks involving the IVC. To provide context for these results, we asked three independent cardiologists to manually assign importance to each view as well as identify whether each structure would be present in each view and compared the model's focus with the consensus cardiologist focus in side-by-side weight matrices (Figure 3D). Both EchoPrime and cardiologists show similar emphasis on important views for different tasks, such as both highlighted the suprasternal notch view for assessing the aorta and the apical views for assessing the left ventricle.

**Transfer Learning and Non-Echocardiographic Disease Prediction**

Finally, we evaluated the performance of EchoPrime's video encoder for both echocardiographic and out-of-domain medical diagnosis tasks to assess whether EchoPrime learns intrinsic knowledge about echocardiography and cardiac pathophysiology. We finetuned vision transformers in a low-data, fully-supervised setting, comparing different initialization weights and found that initializing with EchoPrime weights leads to better performance compared to Kinetics initialization or training from scratch for assessment of mitral regurgitation, identification of pacemaker, diagnosis of cardiac amyloidosis and diagnosis of ST-elevation myocardial infarction (Supplementary Figure 3).

In addition, using EchoPrime's video embeddings, linear probing can identify cardiac diseases not typically diagnosed by echocardiography with high accuracy (Figure 2D) including cardiac amyloidosis and of ST-elevation myocardial infarction. Using EchoPrime embeddings with linear probing, we can identify STEMI with an AUC of 0.90 and amyloidosis with an AUC of 0.95 even with limited training data. It is also possible to predict non-echocardiographic primary diseases without any supervised fine-tuning, with a non-parametric KNN probing approach, that can identify STEMI with an AUC of 0.92, and amyloidosis with an AUC of 0.96

**Discussion**

EchoPrime is a multi-view, view-informed, video-based deep learning algorithm for comprehensive assessment of echocardiograms. Trained on over ten times more data than existing echocardiography AI models, EchoPrime integrates view-dependent information into clinical assessments across a wide range of interpretations of cardiac structure and function. Incorporating videos from more than ten years' worth of echocardiography data at a large academic medical center, EchoPrime integrates anatomic attention and retrieval augmented interpretation to achieve state-of-the art performance on a wide range of interpretation tasks beyond both prior foundation models and task-specific echocardiography AI models. Additionally, EchoPrime excels in many tasks where no prior AI models have been developed and shows generalizability in external validation cohorts. With publicly released code, weights, and demo, we hope EchoPrime can be a resource to the medical and AI communities.

Compared to prior echocardiographic models, EchoPrime has a longer temporal context and integrates complementary videos with distinct information with a robust view classifier and multiple instance attention to integrate the salient features of multiple videos. By utilizing multiple instance learning, we reproduce the clinical relationship between echocardiographic views and associated anatomic structures cardiologists use to make assessments, and it is thus an inherently clinically interpretable model that assesses multiple views and ensembles a wide variety of ultrasound videos. Trained on the largest corpus of echocardiographic data (more than ten times existing echocardiography models), our model can detect rare diseases and cardiovascular diseases not typically assessed by cardiac ultrasound.

A few limitations are worth considering. Further analyses in diverse clinical settings are required. EchoPrime was evaluated on two geographically distinct academic medical centers, and future work should explore its use and utility in diverse settings and more variable image acquisition settings. Point-of-care ultrasound is one high-value use case such that AI can provide rapid expert evaluation of frontline diagnostic information. Additional work is required to continue advancing medical AI models, including the development of multimodal models that integrate complementary diagnostic modalities as well as clinical trials of AI in a clinical setting. Working within a complex ecosystem such as healthcare requires understanding the human-computer interaction and feedback, as well as opportunities to deploy AI models in low-resource settings with out-of-domain or less data.

The Achilles heel of medical imaging lies in human heterogeneity, and opportunities in the future lie in the potential integration of AI into the healthcare system. Our results represent an important step towards the automated evaluation of cardiac ultrasound. EchoPrime augments existing methods for interpreting echocardiography and has potential to streamline manual tasks and improve the reproducibility of cardiac imaging. Trained on the most echocardiography data to date and using anatomic attention as well as retrieval augmented interpretation, EchoPrime performs comprehensive evaluation of echocardiography videos.

## Methods

**Dataset and cohort curation**

The Cedars-Sinai Medical Center (CSMC) echocardiography laboratory performs clinical echocardiography for a wide range of indications ranging from asymptomatic pre-operative screening to evaluation for open heart surgery or heart transplant. A standard full resting echocardiogram study consists of a series of 50-150 videos and images visualizing the heart from different angles and locations, using multiple acquisition techniques including standard 2D videos, tissue Doppler images, and color Doppler videos. Each echocardiogram study corresponds to a unique patient during a specific examination, with different videos representing different facets of cardiac function.

For EchoPrime training, we curated a dataset of 12,124,168 2D and color Doppler echocardiography videos from 275,442 studies and corresponding medical reports from 108,913 patients collected between 2011 and 2022. DICOM images were queried from the clinical data storage system, converted to AVI video files, and de-identified before model training and inference. Data were split by patient into training, validation, and internal test datasets. The training data contained 11,984,170 videos derived from 272,256 studies across 107,663 patients. The validation data contained 25,167 videos from 565 studies across 250 patients, and the internal test set contained 114,831 videos derived from 2,621 studies across 1,000 patients. The full training dataset consists of transthoracic echocardiogram (TTEs) studies, transesophageal (TEEs) echocardiogram and stress echocardiography studies, however downstream experiments and validation tests focused specifically on TTEs, which accounted for 234,036 out of a total of 275,442 studies. For the external test set, we obtained 89,638 videos derived from 1,792 TTE studies and corresponding medical reports across 1792 patients from SHC.

For linear probing and transfer learning experiments, we compiled two focused datasets for diseases not typically diagnosed by echocardiography: one for STEMI and another for amyloidosis. Studies from CSMC were collected and reviewed, with clinical expert labeling of 641 STEMI cases and 159 cardiac amyloidosis cases. To ensure representation in the test set, approximately one-third of the cases were reserved for testing (200 STEMI, 50 amyloidosis), while the remaining cases were used for training.

**Contrastive Visual-Language Training**

To develop video and text encoders that map videos and text into a joint representation space (Figure 1B), we adopted the contrastive strategy used in CLIP[4], but modified the encoder architectures to allow for video and long context text input. For the video encoder, we selected the mVIT[26] architecture with weights pretrained on the Kinetics dataset[27]. The last layer of the video encoder was modified such that the output was 512-dimensional. For the text encoder, we selected Wordpiece Encoding tokenization and BERT[28] architecture with weights pretrained on PubMed abstracts (BioMedBERT[29]). A fully connected layer was appended on top of the BERT classification (CLS) token vector representation to obtain a similarly sized 512-dimensional embedding.

During training, batches of 32 video-report pairs were constructed such that each video and report pair is from a different study. The input video size was 224 x 224 x 16 x 3 with a stride of 2 to increase the temporal context despite only 16 frames. The video data pixel intensity was normalized using the training dataset's mean and standard deviation. For reports longer than 512 tokens, a random 512 tokens were selected starting with either a start token or the first token appearing after a [SEP] token. Videos are fed through the video encoder and reports are fed through the text encoder to obtain embeddings. A dot product is taken between video and report embeddings to obtain a similarity matrix. This similarity matrix is scaled by a trainable temperature parameter (starting at 1.0).

The cross-entropy loss was calculated between this scaled similarity matrix and an identity matrix, which represents the correct matching of video-reports pairs. The AdamW optimizer is used, with a starting learning rate of $4 \times 10^{-5}$, weight decay of $1 \times 10^{-6}$, and the ReduceOnPlateau learning rate scheduler. We froze the weights of the first six layers of the text encoder to retain the general knowledge captured by the pretrained BioMedBERT, while updating all other encoder parameters. The best checkpoint was selected based on the validation loss. With this setup, the model was first pretrained on the full dataset for 60 epochs. It was then fine-tuned on a refined dataset, which included cleaned interpretation text reports that corrected common typographical errors and excluded stress echocardiogram and transesophageal echocardiogram studies for 20 epochs.

**Multiple Instance Learning for Anatomical Attention**

Different echocardiography views offer different insights into cardiac form and function, with some cardiac structures being only visible in certain views but all views providing complementary diagnostic information. We utilized attention-based deep multiple instance learning[30] method to learn relative weights that captures the importance of each echocardiogram view and video for interpretations of different anatomical structures.

We first trained a view classifier to provide a fine-grained characterization of echocardiogram views as defined by the American Society of Echocardiography guidelines[31]. We selected 77,426 echocardiograms for cardiac sonographers to label into 58 different view categories and trained an image-based ConvNextBase[32] view classification model on 224 x 224 images. The AdamW optimizer was employed to minimize the cross-entropy loss, and the best checkpoint was selected based on validation loss. During training random image frames were taken from each labeled video and augmentation techniques including RandAugment[33] and RandomErasing[34] were applied. Held out datasets of studies from different patients were used for validation and testing.

For multiple instance learning (MIL), we use the contrastive video encoder and view classifier to produce a concatenated embedding for each input video of a given echocardiogram study (Supplementary Figure 1). The concatenated embedding is a one hot encoded view classification vector as well as the contrastive video embedding, and MIL is trained to produce importance weights for each video in relation to interpretations separated by anatomic structure. These MIL-generated weights are then used to compute a weighted average of the video embeddings produced by the video encoder. The resulting weighted average as well as individual video level prediction is passed through a multilayer perception to produce an ensembled final prediction. During training, the prediction is compared with the ground truth to compute loss, which is then backpropagated to update the MLP and MIL weights. The output vector of alphas for MIL represent the importance of each view for assessing each anatomic structure. Anatomic structure was defined by structured section in the echocardiogram report ("Left Ventricle", "Right Ventricle", and so on) and each interpretation task was ensembled to create total weighting vectors for each anatomical section, forming the anatomical attention module.

**Cross-Modal Retrieval**

We assess the retrieval accuracy of EchoPrime at the study level. For each echocardiogram study, we generate embeddings for all available videos using the video encoder and averaged to create a study embedding. Similarly, using the text encoder, the report text is mapped to a report text embedding. Using these EchoPrime embeddings, we perform a cross-modal search to find text reports or studies that are semantically similar to a given query study or a query text report by their cosine similarity. To evaluate accuracy, we report the mean and median rank of the correct candidate for both videos-to-report and report-to-videos retrieval. Additionally, we followed the approach used in ALIGN[5] and measured Recall@K, which indicates the percentage of the test set for which the correct result is found within the top K retrieved samples. For comparison, we also calculated retrieval metrics for the previously developed EchoCLIP model.

**Retrieval Augmented Interpretation**

Similar to retrieval augmented generation[18], we leverage the corpus of historical echocardiogram reports and videos for EchoPrime study interpretation. Using pre-trained parametric memory (EchoPrime weights) and non-parametric memory (clinical reports), our approach leverages anatomic attention to weight the cross-modal retrieval based on each section in an echocardiogram report. Interpretations are weighted by anatomy and video to obtain comprehensive echocardiography exam interpretations. The anatomical attention module identifies the most informative views for the specified section and computes a weighted average of video embeddings based on their views, resulting in a section-specific study embedding. We use this method to generate 50 candidate interpretation reports and average features across these 50 reports to obtain final predictions.

For example, in determining the left ventricular ejection fraction (LVEF), anatomical attention weights video embeddings (apical-4-chamber, apical-2-chamber for example) important for assessing the LV to produce a unified anatomical embedding. Each video provides different information—an apical-4-chamber view might predict an LVEF of 34%, while the apical-2-chamber view might predict 38%. However, by synthesizing these embeddings into a unified anatomical embedding, we can extract the relevant information from each video and combine them to make a more accurate prediction. This approach does not require additional fine-tuning and can generalize to predict any feature present in the reports.

**Disease Diagnosis with Probing**

We apply k-nearest neighbors (KNN) and linear probing on top of EchoPrime for non-echocardiographic disease prediction. STEMI and amyloidosis datasets are used for training and validation of probing approaches. Study embeddings are obtained by averaging the embeddings of all videos within each study, and used along with clinician-assigned diagnoses to fit the models. For KNN probing, we use KNeighborsClassifier, and for linear probing, we apply LogisticRegression, both from scikit-learn library.

**Model Performance with Different Initialization**

We test whether initializing the model with EchoPrime weights leads to higher label efficiency and accuracy compared to initializing with Kinetics weights and random initialization. We selected four binary classification tasks: pacemaker detection, mitral regurgitation classification, amyloidosis detection, and STEMI detection. Pacemaker detection was trained on A4C RV views, mitral regurgitation on Doppler A4C MV views, amyloid detection on PLAX views, and STEMI detection was trained on A4C views. All tasks are trained on datasets of 32, 64, 128, and 256 samples, with an equal distribuion of positive and negative samples.

**Evaluation and Benchmarking**

To evaluate the accuracy of echocardiography interpretation, we chose 19 cardiac features of different anatomic structures, including 17 binary and 2 continuous features (Supplementary Table 1). We identified corresponding report phrases in report text to obtain ground truth labels based on the presence or absence of a specific interpretation or numerical measurements. For tasks assessing severity (trivial, mild, moderate, severe), we binarized assessments by considering a feature as present if the severity is classified as moderate or severe. Accuracy for classification tasks was measured with the area under the receiver operating characteristic curve (AUROC) and coefficient of determination (R2 score) for regression tasks. To assess the accuracy of our multi-label view classifier, we employed the average AUROC using a One-vs-Rest approach, treating the correct class as the positive label and all other classes as the negative label and averaged the results across all classes.

To compare the performance with other foundation models, we also obtained interpretations using EchoCLIP and BioMedCLIP using retrieval. We selected one A4C video per study, embedded each frame of the video, and averaged the embedding to produce a study embedding. Then, similarly to EchoPrime retrieval, we found the 50 closest reports to this study embedding and averaged their features to make final predictions. We used this method of retrieval because while both EchoCLIP and BioMedCLIP are image-based models, A4C is considered the most informative view by physicians (Figure 3D) and this approach follows the methodology outlined in the EchoCLIP paper.

**Computing Hardware and Software**

We used Python (v3.8.13) and PyTorch (v2.1.2, CUDA 12.1) (https://pytorch.org) for all experiments and analyses in the study. We used scikit-learn (https://scikit-learn.org/) for probing methods and umap-learn (https://umap-learn.readthedocs.io) for dimensionality reduction. To train EchoPrime we used two 50-GB NVIDIA RTX A6000 GPUs configured for multi-GPU training using PyTorch's DistributedDataParallel. We obtained weights of previously developed foundation models from Hugging Face model hub ( https://huggingface.co/docs/hub/en/models-the-hub ): EchoCLIP ( https://huggingface.co/mkaichristensen/echo-clip-r ), BioMedCLIP ( https://huggingface.co/microsoft/BiomedCLIP-PubMedBERT_256-vit_base_patch16_224 ).

**Data availability**

The dataset of videos and reports used to train EchoCLIP is not publicly available due to its potentially identifiable nature.

**Code Availability**

The code for EchoPrime is available at https://github.com/echonet/EchoPrime.

**Acknowledgements**

This work is funded by NIH NHLBI grants R00HL157421, R01HL173526, and R01HL173487 to DO.

## References

1. Liu, X. *et al.* A comparison of deep learning performance against health-care professionals in detecting diseases from medical imaging: a systematic review and meta-analysis. *Lancet Digit. Health* **1**, e271–e297 (2019).
2. Esteva, A. *et al.* Dermatologist-level classification of skin cancer with deep neural networks. *Nature* **542**, 115–118 (2017).
3. McKinney, S. M. *et al.* International evaluation of an AI system for breast cancer screening. *Nature* **577**, 89–94 (2020).
4. Radford, A. *et al.* Learning Transferable Visual Models From Natural Language Supervision. in *Proceedings of the 38th International Conference on Machine Learning* 8748–8763 (PMLR, 2021).
5. Jia, C. *et al.* Scaling Up Visual and Vision-Language Representation Learning With Noisy Text Supervision. in *Proceedings of the 38th International Conference on Machine Learning* 4904–4916 (PMLR, 2021).
6. Li, J., Li, D., Xiong, C. & Hoi, S. BLIP: Bootstrapping Language-Image Pre-training for Unified Vision-Language Understanding and Generation. in *Proceedings of the 39th International Conference on Machine Learning* 12888–12900 (PMLR, 2022).
7. Singh, A. *et al.* FLAVA: A Foundational Language And Vision Alignment Model. in *2022 IEEE/CVF Conference on Computer Vision and Pattern Recognition (CVPR)* 15617–15629 (2022). doi:10.1109/CVPR52688.2022.01519.
8. Li, H. *et al.* Uni-Perceiver v2: A Generalist Model for Large-Scale Vision and Vision-Language Tasks. in *2023 IEEE/CVF Conference on Computer Vision and Pattern Recognition (CVPR)* 2691–2700 (IEEE, Vancouver, BC, Canada, 2023). doi:10.1109/CVPR52729.2023.00264.
9. Alayrac, J.-B. *et al.* Flamingo: a Visual Language Model for Few-Shot Learning. *Adv. Neural Inf. Process. Syst.* **35**, 23716–23736 (2022).
10. Wang, W. *et al.* Image as a Foreign Language: BEIT Pretraining for Vision and Vision-Language Tasks. in *2023 IEEE/CVF Conference on Computer Vision and Pattern Recognition (CVPR)* 19175–19186 (IEEE, Vancouver, BC, Canada, 2023). doi:10.1109/CVPR52729.2023.01838.
11. Zhai, X., Mustafa, B., Kolesnikov, A. & Beyer, L. Sigmoid Loss for Language Image Pre-Training. in *2023 IEEE/CVF International Conference on Computer Vision (ICCV)* 11941–11952 (IEEE, Paris, France, 2023). doi:10.1109/ICCV51070.2023.01100.
12. Lavoie, S. *et al.* Modeling Caption Diversity in Contrastive Vision-Language Pretraining. Preprint at http://arxiv.org/abs/2405.00740 (2024).
13. Chen, R. J. *et al.* Towards a general-purpose foundation model for computational pathology. *Nat. Med.* **30**, 850–862 (2024).
14. Huang, K. *et al.* A foundation model for clinician-centered drug repurposing. *Nat. Med.* 1–13 (2024) doi:10.1038/s41591-024-03233-x.
15. Bannur, S. *et al.* Learning to Exploit Temporal Structure for Biomedical Vision-Language Processing. in *2023 IEEE/CVF Conference on Computer Vision and Pattern Recognition (CVPR)* 15016–15027 (IEEE, Vancouver, BC, Canada, 2023). doi:10.1109/CVPR52729.2023.01442.
16. Zhou, Y. *et al.* A foundation model for generalizable disease detection from retinal images. *Nature* 1–8 (2023) doi:10.1038/s41586-023-06555-x.

17. Christensen, M., Vukadinovic, M., Yuan, N. & Ouyang, D. Vision–language foundation model for echocardiogram interpretation. *Nat. Med.* **30**, 1481–1488 (2024).
18. Lewis, P. *et al.* Retrieval-Augmented Generation for Knowledge-Intensive NLP Tasks. in *Advances in Neural Information Processing Systems* vol. 33 9459–9474 (Curran Associates, Inc., 2020).
19. Ouyang, D. *et al.* Video-based AI for beat-to-beat assessment of cardiac function. *Nature* **580**, 252–256 (2020).
20. Vrudhula, A. *et al.* Deep Learning Phenotyping of Tricuspid Regurgitation for Automated High Throughput Assessment of Transthoracic Echocardiography. *medRxiv* 2024.06.22.24309332 (2024) doi:10.1101/2024.06.22.24309332.
21. Vrudhula, A. *et al.* High-Throughput Deep Learning Detection of Mitral Regurgitation. *Circulation* **150**, 923–933 (2024).
22. Yıldız Potter, İ., Leo, M. M., Vaziri, A. & Feldman, J. A. Automated detection and localization of pericardial effusion from point-of-care cardiac ultrasound examination. *Med. Biol. Eng. Comput.* **61**, 1947–1959 (2023).
23. Holste, G. *et al.* Severe aortic stenosis detection by deep learning applied to echocardiography. *Eur. Heart J.* **44**, 4592–4604 (2023).
24. Ghorbani, A. *et al.* Deep learning interpretation of echocardiograms. *Npj Digit. Med.* **3**, 1–10 (2020).
25. Zhang, S. *et al.* BiomedCLIP: a multimodal biomedical foundation model pretrained from fifteen million scientific image-text pairs. Preprint at https://doi.org/10.48550/arXiv.2303.00915 (2024).
26. Fan, H. *et al.* Multiscale Vision Transformers. in *Proceedings of the IEEE/CVF International Conference on Computer Vision* 6824–6835 (2021).
27. Kay, W. *et al.* The Kinetics Human Action Video Dataset. Preprint at http://arxiv.org/abs/1705.06950 (2017).
28. Devlin, J., Chang, M.-W., Lee, K. & Toutanova, K. BERT: Pre-training of Deep Bidirectional Transformers for Language Understanding. in *Proceedings of the 2019 Conference of the North American Chapter of the Association for Computational Linguistics: Human Language Technologies, Volume 1 (Long and Short Papers)* (eds. Burstein, J., Doran, C. & Solorio, T.) 4171–4186 (Association for Computational Linguistics, Minneapolis, Minnesota, 2019). doi:10.18653/v1/N19-1423.
29. Gu, Y. *et al.* Domain-Specific Language Model Pretraining for Biomedical Natural Language Processing. *ACM Trans. Comput. Healthc.* **3**, 1–23 (2022).
30. Ilse, M., Tomczak, J. & Welling, M. Attention-based Deep Multiple Instance Learning. in *Proceedings of the 35th International Conference on Machine Learning* 2127–2136 (PMLR, 2018).
31. Mitchell, C. *et al.* Guidelines for Performing a Comprehensive Transthoracic Echocardiographic Examination in Adults: Recommendations from the American Society of Echocardiography. *J. Am. Soc. Echocardiogr.* **32**, 1–64 (2019).
32. Liu, Z. *et al.* A ConvNet for the 2020s. in *Proceedings of the IEEE/CVF Conference on Computer Vision and Pattern Recognition* 11976–11986 (2022).
33. Cubuk, E. D., Zoph, B., Shlens, J. & Le, Q. RandAugment: Practical Automated Data Augmentation with a Reduced Search Space. in *Advances in Neural Information Processing Systems* vol. 33 18613–18624 (Curran Associates, Inc., 2020).

34. Zhong, Z., Zheng, L., Kang, G., Li, S. & Yang, Y. Random Erasing Data Augmentation. *Proc. AAAI Conf. Artif. Intell.* **34**, 13001–13008 (2020).

**Table 1:** Clinical characteristics of the study cohorts.

|  | Cedars Sinai | Stanford |
|---|---:|---:|
| **Videos** | 12,124,168 | 91,746 |
| **Studies (N)** | 275,442 | 1,792 |
| **Patients** | 108,913 | 1,779 |
| **Age (mean (s.d.))** | 66.31 (16.7) | 60.42 (17.3) |
| **Female (%)** | 117,324 (42.6) | 843 (47.0) |
| **Race (%)** | | |
| Non-Hispanic White | 166,091 (60.3) | 900 (50.2) |
| Black | 35,624 (12.9) | 76 (4.2) |
| Hispanic | 27,194 (9.9) | 241 (13.5) |
| Asian | 20,299 (7.4) | 301 (16.8) |
| Other | 19,832 (7.2) | 161 (9.0) |
| Unknown | 5,321 (1.9) | 85 (4.7) |
| Pacific Islander | 984 (0.4) | 28 (1.6) |
| **Conditions (%)** | | |
| Hypertension | 112,919 (41.0) | 1055 (59.9) |
| Heart failure | 93,875 (34.1) | 746 (41.6) |
| Atrial fibrillation | 60,448 (21.9) | 555 (30.0) |
| Chronic kidney disease | 50,302 (18.3) | 580 (32.4) |
| Diabetes mellitus | 46,656 (16.9) | 168 (9.3) |
| Pulmonary artery disease | 28,903 (10.5) | 41 (2.3) |
| Cerebrovascular accident | 18,442 (6.7) | 104 (5.8) |
| Myocardial infarction | 18,283 (6.6) | 296 (16.5) |

Table 2: Echocardiography interpretation performance with benchmarks of other foundation models and task specific models. For classification tasks, we report AUROC and regression tasks are evaluated using R2 and mean square error (MAE). The best is bolded. LV = Left Ventricle. RV = Right Ventricle. LA = Left Atrium. RA = Right Atrium. TAVR – Transcatherer Aortic Valve Replacement. AV – Aortic Valve. IVC – Inferior Vena Cava. PA – Pulmonary Artery.

| | INTERNAL TEST | | | | EXTERNAL TEST | | | |
|---|---|---|---|---|---|---|---|---|
| | BioMed CLIP | Echo CLIP | Echo Prime | Task specific model | BioMed CLIP | Echo CLIP | Echo Prime | Task specific model |
| **LV ejection fraction ($R^2$)** | -2.60 | 0.62 | **0.83** | 0.81[19] | -3.50 | 0.48 | **0.79** | 0.77[19] |
| **LV ejection fraction (MAE)** | 26.93 | 7.00 | 4.79 | **4.10** | 23.00 | 6.23 | **4.14** | 6.00 |
| **Pacemaker** | 0.58 | **0.88** | 0.85 | - | 0.64 | **0.90** | 0.84 | - |
| **RV systolic function** | 0.64 | 0.89 | **0.93** | - | 0.59 | 0.89 | **0.94** | - |
| **RV dilation** | 0.58 | 0.88 | **0.89** | - | 0.53 | 0.84 | **0.85** | - |
| **LA dilation** | 0.66 | **0.93** | 0.91 | 0.85[24] | 0.57 | 0.67 | **0.73** | - |
| **RA dilation** | 0.67 | 0.90 | **0.90** | - | 0.56 | 0.72 | **0.77** | - |
| **Mitraclip** | 0.70 | 0.95 | **0.99** | - | 0.79 | 0.81 | **0.98** | - |
| **Mitral annular calcification** | 0.63 | 0.95 | **0.96** | - | 0.66 | 0.94 | **0.96** | - |
| **Mitral stenosis** | 0.70 | **0.96** | 0.96 | - | 0.72 | 0.84 | **0.92** | - |
| **Mitral regurgitation** | 0.66 | 0.89 | **0.92** | 0.92[21] | 0.67 | 0.86 | 0.91 | **0.95**[21] |
| **TAVR** | 0.54 | 0.86 | **1.00** | - | 0.52 | 0.95 | **0.97** | - |
| **Bicuspid AV** | 0.52 | 0.72 | **0.83** | - | 0.50 | 0.63 | **0.82** | - |
| **Aortic stenosis** | 0.55 | 0.83 | **0.98** | 0.98[23] | 0.52 | 0.81 | **0.96** | 0.95[23] |
| **Aortic regurgitation** | 0.54 | 0.68 | **0.88** | - | 0.53 | 0.67 | **0.89** | - |
| **Tricuspid regurgitation** | 0.69 | 0.91 | **0.95** | 0.93[20] | 0.61 | 0.80 | 0.88 | **0.95**[20] |
| **Pericardial effusion** | 0.68 | 0.92 | **0.98** | 0.94[22] | 0.75 | 0.89 | **0.89** | - |
| **Aortic root dilation** | 0.50 | 0.76 | **0.91** | - | 0.52 | 0.71 | **0.97** | - |
| **Dilated IVC** | 0.49 | 0.81 | **0.84** | - | 0.46 | 0.75 | **0.86** | - |
| **PA pressure ($R^2$)** | -0.19 | 0.35 | **0.43** | - | 0.25 | 0.22 | **0.36** | - |
| **PA pressure (MAE)** | 10.62 | 7.84 | **7.30** | - | 10.92 | 8.80 | **7.97** | - |

Table 3: Cross-modal retrieval metrics on a test set of 2,254 videos-text report pairs. Lower rank is better, and higher recall is better. Best performance is bolded.

|  | Videos-To-Text | | | | | Text-To-Videos | | | | |
| --- | --- | --- | --- | --- | --- | --- | --- | --- | --- | --- |
|  | Mean Rank | Median Rank | Recall @1 | Recall @5 | Recall @10 | Mean Rank | Median Rank | Recall @1 | Recall @5 | Recall @10 |
| **EchoPrime** | 2.86 | 1 | 70% | 94% | 98% | 3.05 | 1 | 69% | 94% | 97% |
| **EchoCLIP** | 58.71 | 9 | 20% | 42% | 53% | 37.07 | 5 | 25% | 50% | 62% |

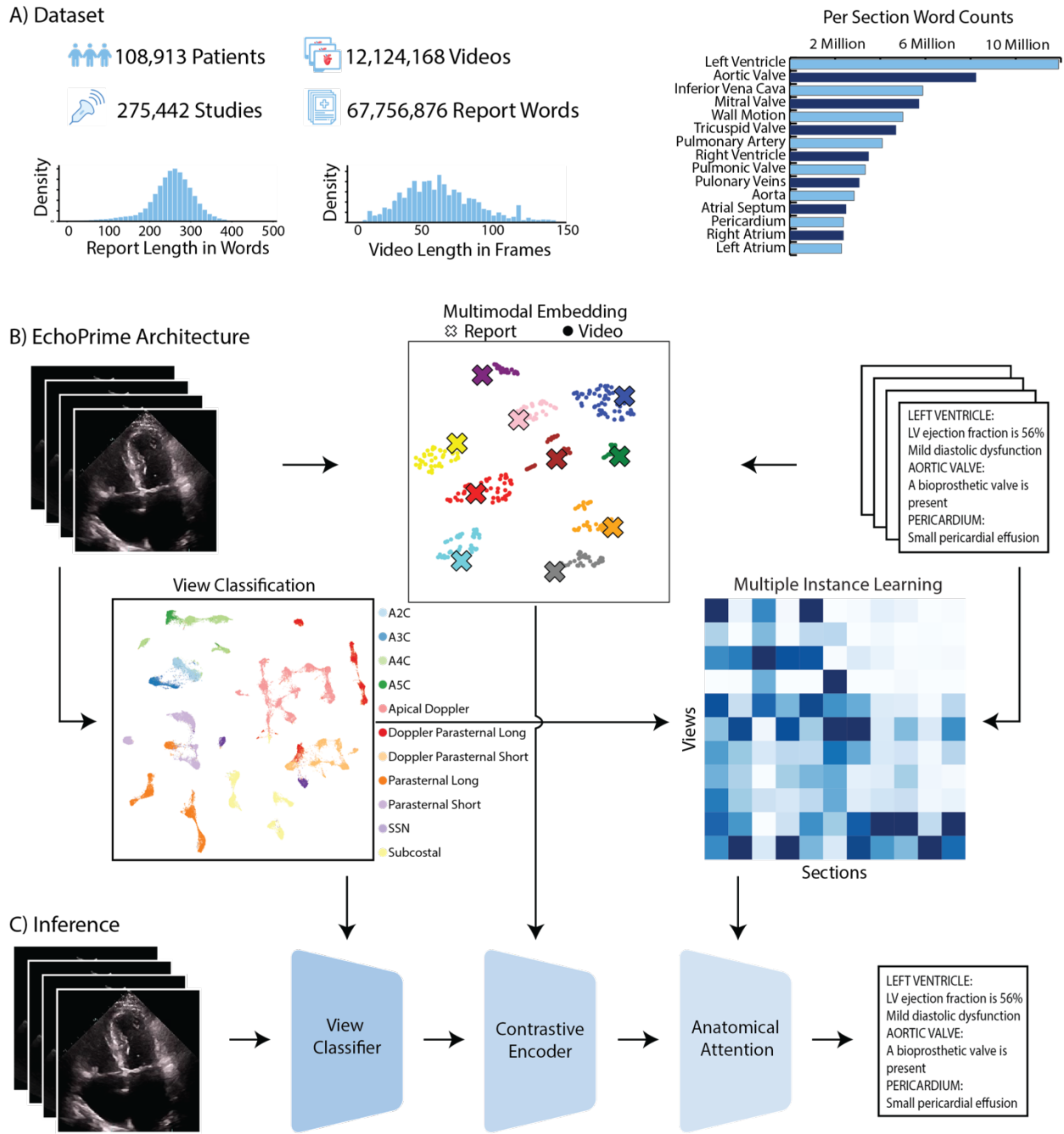

**Figure 1 | EchoPrime Overview. (A)** Training Dataset Characteristics. **(B)** Video and text encoders are trained to map videos and text into a joint latent space. The view classifier and multiple instance learning are trained separately to learn view weighting based on the anatomical section. **(C)** Inference Pipeline: When given a comprehensive echocardiogram study, EchoPrime determines view for each video, uses contrastive encoder on joint video-text latent space for

retrieval augmented interpretation, and weights the interpretation of each video by anatomical attention for a study-level assessment of each cardiac structure.

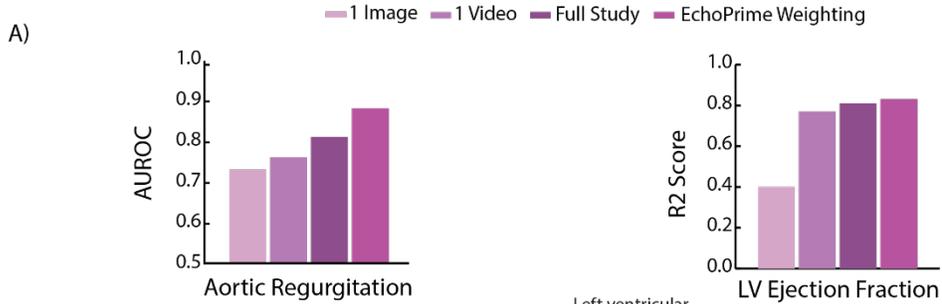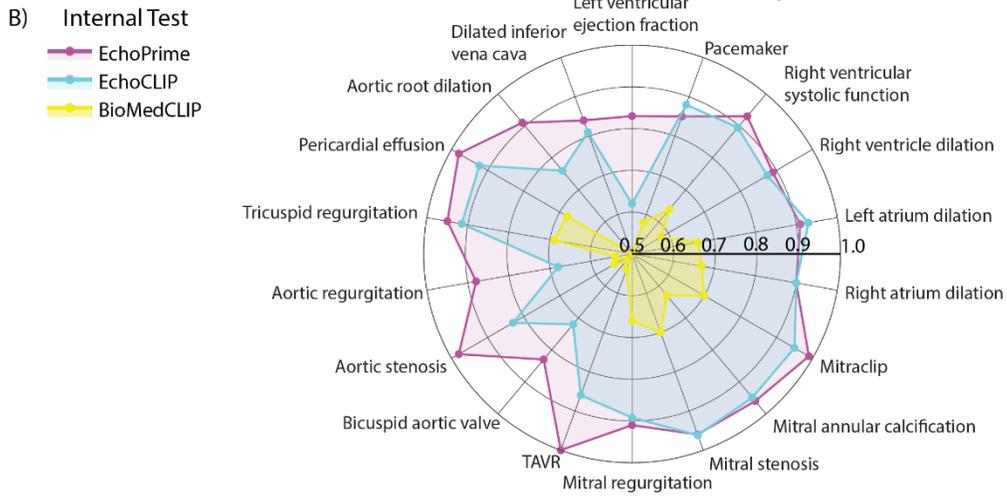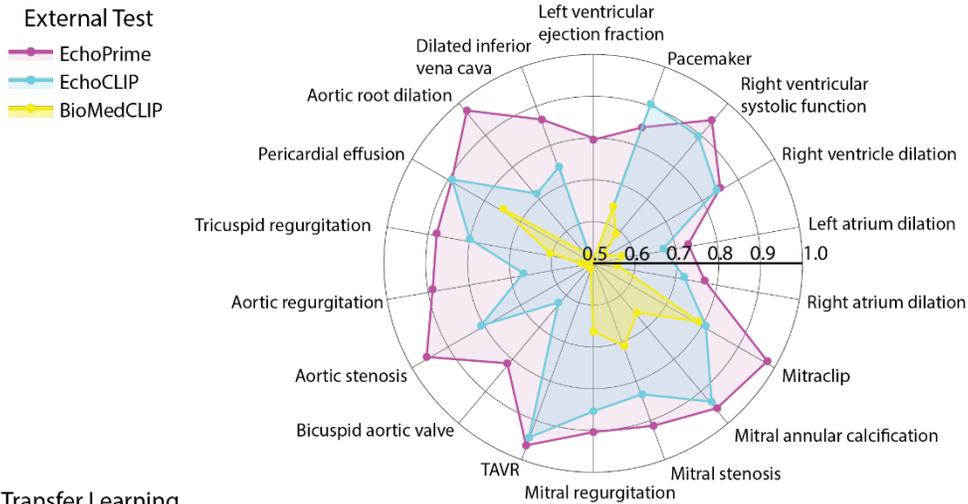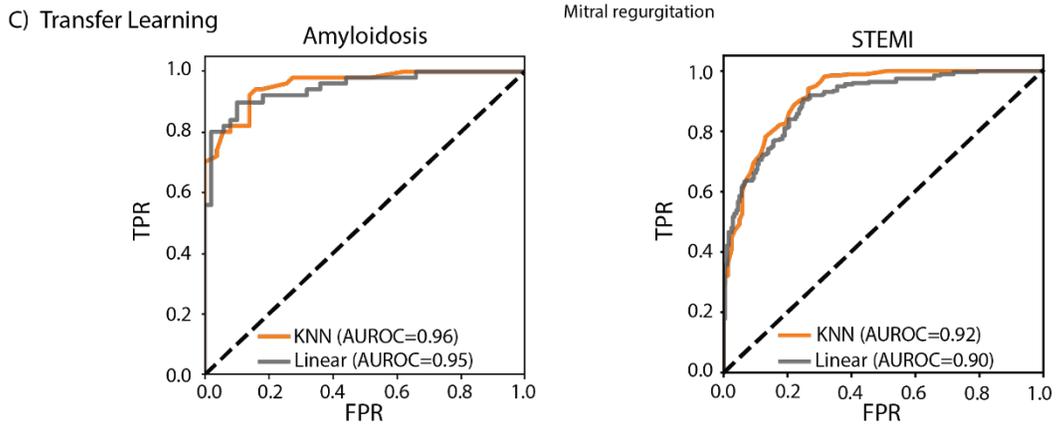

**Figure 2 | Performance metrics for echocardiography prediction tasks. (A)** The accuracy of predicting aortic regurgitation severity and left ventricular ejection fraction improves progressively from predictions on a single image to a full video, a single video to multiple videos in a full study, and finally applying anatomical attention (EchoPrime weighting). **(B)** Comparison of EchoPrime, EchoCLIP, and BioMedCLIP on a range of echocardiography interpretation tasks on internal and external test cohorts. **(C)** Using linear probing techniques to assess transfer learning, EchoPrime accurately predicts diseases even when their labels are not explicitly mentioned in the reports (non-echocardiographic diseases).

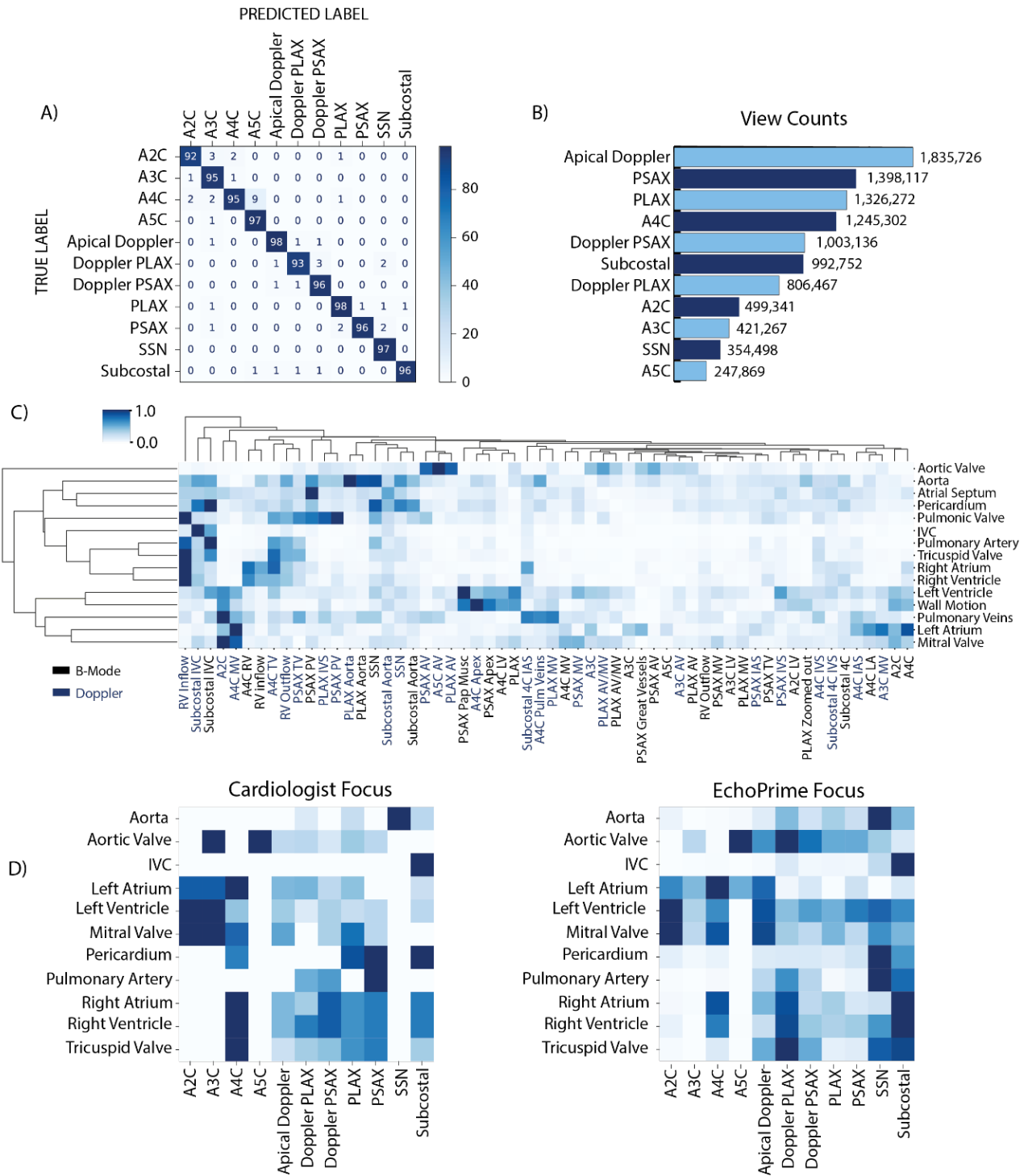

**Figure 3 Anatomic Attention to weight predictions across videos in a echocardiogram study**
**(A)** A view classifier was trained on 60k videos to distinguish between 58 standard echocardiographic views. **(B)** A comprehensive echocardiogram exam comprises of many different views, which is summarized across our training cohort. **(C)** A clustered heatmap showing the relative priority and ranking of each video for each anatomic structure based on learned anatomical attention **(D)** Comparison of how a cardiologist (left) and EchoPrime (right) prioritizes different views based on the assessed anatomic structure.